\documentclass[12pt]{spieman}  
\usepackage{amsmath,amsfonts,amssymb}
\usepackage{graphicx}
\usepackage{setspace}
\usepackage{tocloft}
 
\usepackage{amsmath,amsfonts,amssymb}
\usepackage{graphicx}
\usepackage{algorithm}
\usepackage[noend]{algpseudocode}
\usepackage{verbatim}
\usepackage[shortlabels]{enumitem}
\usepackage{soul}
\usepackage{forest}
\usepackage{mathrsfs}
\usepackage{booktabs}
\usepackage{amssymb}
\usepackage{makecell}
\usepackage{multirow}
\usepackage{graphicx}
\usepackage{rotating}
\usepackage{color,colortbl}
\usepackage{comment}
\usepackage{lineno}
\def\CC{{C\nolinebreak[4]\hspace{-.05em}\raisebox{.4ex}{\tiny\bf ++}}}

\title{Machine learning for automatic construction of  pseudo-realistic pediatric abdominal phantoms}

\author[a,*]{Marco Virgolin}
\author[b]{Ziyuan Wang}
\author[b]{Tanja Alderliesten}
\author[a,c]{Peter A. N. Bosman}
\affil[a]{Life Sciences and Health Group, Centrum Wiskunde \& Informatica, Science Park 123, Amsterdam, the Netherlands}
\affil[b]{Department of Radiation Oncology, Amsterdam UMC, University of Amsterdam, Meibergdreef 9, Amsterdam, the Netherlands}
\affil[c]{Algorithmics Group, Delft University of Technology, Mekelweg 5, Delft, the Netherlands}

\cftpagenumbersoff{figure}
\cftpagenumbersoff{table} 
\begin{document} 
\maketitle

\begin{abstract}
Machine Learning (ML) is proving extremely beneficial in many healthcare applications. In pediatric oncology, retrospective studies that investigate the relationship between treatment and late adverse effects still rely on simple heuristics. To assess the effects of radiation therapy, treatment plans are typically simulated on phantoms, i.e., virtual surrogates of patient anatomy. Currently, phantoms are built according to reasonable, yet simple, human-designed criteria. This often results in a lack of individualization. We present a novel approach that combines imaging and ML to build individualized phantoms automatically. Given the features of a patient treated historically (only 2D radiographs available), and a database of 3D Computed Tomography (CT) imaging with organ segmentations and relative patient features, our approach uses ML to predict how to assemble a patient-specific phantom automatically. Experiments on 60 abdominal CTs of pediatric patients show that our approach constructs significantly more representative phantoms than using current phantom building criteria, in terms of location and shape of the abdomen and of two considered organs, the liver and the spleen. Among several ML algorithms considered, the Gene-pool Optimal Mixing Evolutionary Algorithm for Genetic Programming (GP-GOMEA) is found to deliver the best performing models, which are, moreover, transparent and interpretable mathematical expressions.
\end{abstract}


{\noindent \footnotesize\textbf{*}Marco Virgolin,  \linkable{marco.virgolin@cwi.nl} }



\section{Introduction}
Virtual anthropomorphic phantoms are 3D representations of the human body that are used as surrogates for the anatomy of humans, to estimate the quantity and geometric distribution of radiation dose when having been exposed to radiation, e.g., in radiation treatment for cancer patients~\cite{lee2015reconstruction,xu2014exponential}. 
Because anatomy resemblance is one of the key sources of uncertainty in dose estimation~\cite{bezin2017review}, the phantom needs to represent the anatomy of the patient for whom estimates are needed with high precision.

Current methods for phantom building have two major limitations. Firstly, traditionally, building phantoms is a manual and time-consuming task. Approximations of human anatomies are produced using simple geometrical shapes~\cite{stovall2004genetic,howell2019adaptations}, or by considering actual organ segmentations from Computed Tomography (CT) scans~\cite{geyer2014uf,alziar2009individual,xie2017computational}. These anatomies are shaped and/or adapted according to population-based statistics and/or reasonable human-designed criteria~\cite{valentin2002basic,kuczmarski1994increasing,stovall2004genetic,geyer2014uf,alziar2009individual}. Because the procedure is laborious, a limited number of phantoms is made, each meant to represent a category of patients.
The second and perhaps more fundamental limitation is that it is unknown how to best define categorization criteria that best capture resemblance in individual patient anatomy. So far, only simple criteria such as partitioning by combinations of age, gender, percentiles height and weight, have been explored~\cite{stovall2004genetic,geyer2014uf,alziar2009individual,xie2017computational}. Nevertheless, several studies have indicated that such simple criteria are incapable of capturing the high variance in human internal anatomy~\cite{xu2014exponential,geyer2014uf,de2001organ,virgolin2018feasibility}, and this can ultimately lead to coarse dose estimations~\cite{wang2018age}.

Machine Learning (ML) is becoming more and more a reliable approach to tackle hard and heterogeneous problems in healthcare~\cite{obermeyer2016predicting}. This is because ML can infer patterns from data that are hard to spot, and to model for humans. In the context of phantom construction, the use of ML could improve upon the rough categorization methods that are currently being employed.
In this article, we present a new take on phantom construction that shows that it is possible to use ML to obtain better phantoms. 
In particular, we propose an automatic phantom-construction pipeline that can be used to generate pseudo-realistic phantoms that are patient-specific. To overcome the need for laborious manual intervention, we propose to re-use 3D patient imaging (CT scans and organ segmentations) collected in a database, to assemble new anatomy combinations. To estimate how to best perform this assembling, i.e., to move beyond the use of too simplistic criteria, we rely on ML. Specifically, we train ML models to learn relationships between patient features and 3D metrics based on their internal anatomy. 

We consider a relatively hard scenario where phantoms are needed and patient features are limited: dose reconstruction for historical patients, i.e., patients treated in the pre-3D planning era, when radiation treatment plans were designed using 2D radiographs. As no 3D imaging is available for historical patients to simulate the treatment and estimate the radiation dose distribution, phantoms are necessary to act as surrogate anatomies in order to reconstruct 3D dose distributions~\cite{virgolin2018feasibility,wang2018age}.
We focus on children between 2 to 6 years, and on dose reconstruction for abdominal radiation treatment, for the following reasons. Firstly, children are typically under-represented in existing phantom libraries, i.e., phantoms are available for few categories~\cite{stovall2004genetic,geyer2014uf}. Secondly, the inclusion of radiation treatment has led to high survival rates for several types of pediatric abdominal cancer (e.g., Wilms' tumor, the most common type of kidney cancer), but it is known to cause late adverse effects~\cite{breslow1993epidemiology,van2010evaluation}.
Thirdly, it has recently been shown that when CT scans are selected based on age and gender to serve as a surrogate for pediatric abdominal patients, there is a high risk of obtaining inaccurate dose reconstructions~\cite{wang2018age}. 
The ultimate goal is to realize sufficiently accurate dose reconstruction by use of more representative phantoms, which can then be used to better understand how radiation dose contributes to the onset of late adverse effects. Providing this information can support radiation oncologists in the design of better treatment plans with smaller chances of adverse effects for today's abdominal radiation treatment.



\section{Materials and Methods}

\subsection{Data}\label{sec:data}
We built a database using data of 60 pediatric cancer patients, in the age range of 2 to 6 years. The patients were treated after 2002 at the radiation oncology department of the Amsterdam UMC, location AMC, in Amsterdam, or at the University Medical Center Utrecht/Princess M\`axima Center for Pediatric Oncology in Utrecht. For each patient, a CT is available that fully includes the lower part of the thorax to the lower part of the abdomen, specifically from the top of the Thoracic 10th vertebra (T10) to the bottom of the Sacral 1st vertebra (S1). The median axial thickness of the CT scan is 2.5mm, and the median in-plane resolution is $1.0\text{mm}\times1.0$mm.

\begin{figure}
\centering
\begin{tabular}{c c c}
\includegraphics[width=.27\linewidth]{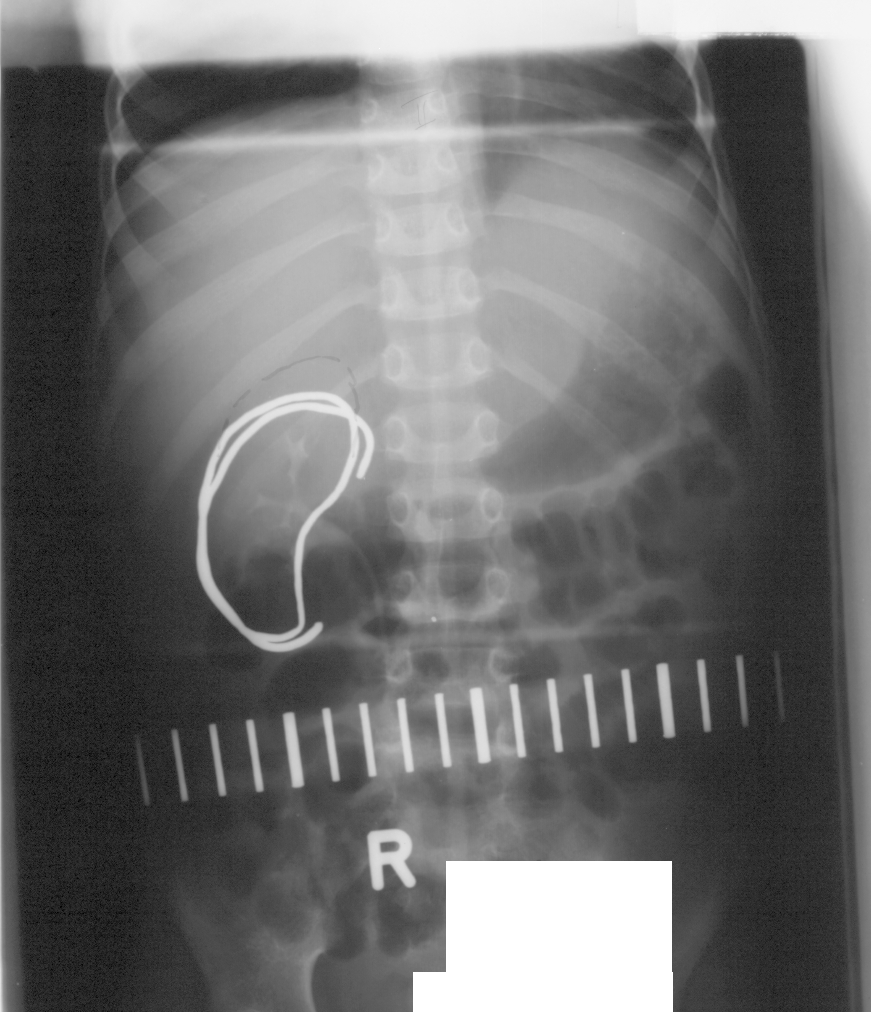} &
\includegraphics[width=.27\linewidth]{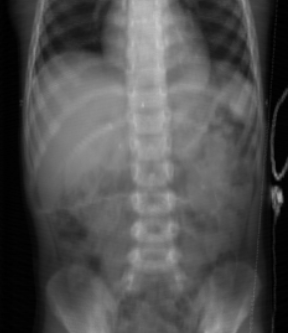} &
\includegraphics[width=.27\linewidth]{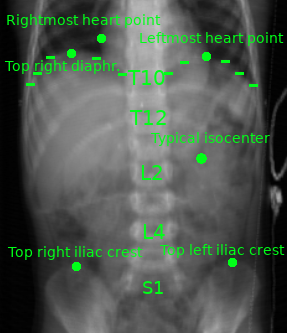}
\\
\end{tabular}
\caption{Left: An example of a 2D coronal radiograph, taken by a radiation treatment simulator used in the pre-3D planning era, including annotations by medical personnel (sensitive information censored). Middle: A digitally reconstructed radiograph built from a CT. Images are acquired in anterior-posterior setting. Liver and spleen are not clearly visible. Right: Example of manually-placed landmarks used to measure features from radiographs. The length of the left and the right diaphragm along the LR direction is derived by fitting a cubic spline to the respective dashes.}
\label{fig:radiograph}
\end{figure}

To simulate the scenario of dose reconstruction for patients treated in the pre-3D planning era, we only consider patient features that were typically recorded at the time. We base our choices on the availability of features for the Emma Children's Hospital/Academic Medical Center (EKZ/AMC) childhood cancer survivor cohort, treated between 1966 and 1996~\cite{van2010evaluation}. One important source of information for the EKZ/AMC cohort is 2D coronal radiographs, which were acquired to plan radiation treatment. The coronal radiographs enable us to perform measurements, along Left-Right (LR) and Inferior-Superior (IS) directions, based on visible anatomical landmarks, i.e., the bony anatomy (see Fig.~\ref{fig:radiograph}). For our cohort of recently treated patients, we simulate historical radiographs using Digitally Reconstructed Radiographs (DRRs), derived from the CTs using in-house developed software. Figure~\ref{fig:radiograph} (left, middle) shows an example of a historical radiograph and of a DRR. 

\begin{figure}[h]
\centering
\includegraphics[width=0.6\linewidth]{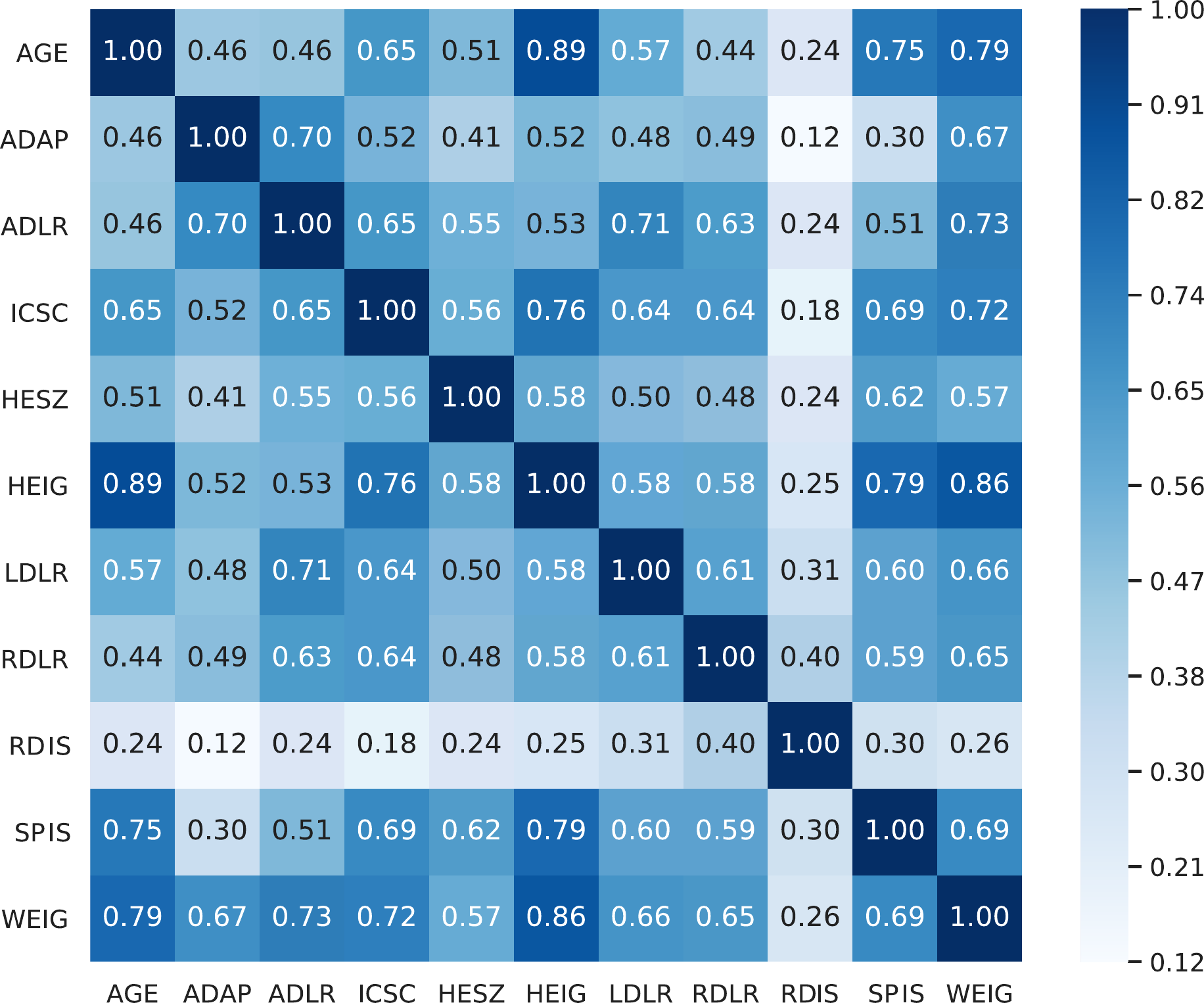}
\caption{Pearson correlation coefficients (number and color-coded) between the considered features. See Table~\ref{tab:features} for the meaning of abbreviations.}
\label{fig:corrmat}
\end{figure}

Table~\ref{tab:features} lists the features considered in this work. Features involving measurements from DRRs were collected after manual placement of landmarks (using 3D Slicer software\cite{fedorov20123d}), exemplified in Fig.~\ref{fig:radiograph} (right).
The only source of information on the Anterior-Posterior (AP) direction is the abdominal diameter, that was historically measured using rulers and calipers, at the center of the radiation treatment field (which corresponded with the isocenter for the EKZ/AMC cohort). For our cohort, we measured the abdominal diameter along AP from the CTs, using a typical isocenter position for abdominal flank irradiation, as described in our previous work~\cite{virgolin2018feasibility}.
Figure~\ref{fig:corrmat} shows the Pearson correlation coefficients between the considered features. Most features are moderately correlated, and few are strongly correlated, e.g., height with age and weight. The distance along IS between the top of the right diaphragm and T12 (RDIS) stands out as it is associated with the lowest correlations with any other feature. We measured this feature in an attempt to capture information on the breathing state of the patient, which is known to be correlated with organ position~\cite{huijskens2017magnitude,xi2009analysis}. The low correlation is likely because the particular breathing state of the patient can be reasonably expected to be not correlated with the other features we considered.

\begin{table}
\caption{Features of our cohort, typically available for patients treated in the pre-3D planning era. Note: gender is categorical, other features are numerical.}
\label{tab:features}
\centering
\small
\scalebox{0.8}{
\begin{tabular}{l c c c c c c c}
\toprule
Feature name & Abbreviation & Unit & Source & Min & Max & Mean & St.Dev.\\
\midrule
Age & AGE & years & records & 2.0 & 6.0 & 3.8 & 1.2\\
Abdominal diameter in AP at typical isocenter & ADAP & mm & records & 11.1 & 16.0 & 13.3 & 1.2\\
Abdominal diameter in LR at middle of L2 & ADLR & mm & radiograph & 16.3 & 23.5 & 19.4 & 1.4\\
Distance from top of iliac crest to spinal cord along LR & ICSC & mm & radiograph & 4.3 & 6.8 & 5.5 & 0.6\\
Gender & GEND & -- & records & \multicolumn{4}{c}{33 females, 27 males} \\
Heart size along LR & HESZ & mm & radiograph & 6.8 & 9.9 & 8.5 & 0.7\\
Height & HEIG & cm & records & 86.0 & 123.0 & 103.0 & 10.7\\
Left diaphragm length along LR & LDLR & mm & radiograph & 6.5 & 10.7 & 8.4 & 0.9\\
Right diaphragm length along LR & RDLR & mm & radiograph & 6.2 & 10.5 & 8.3 & 0.8\\
Right diaphragm top to T12 distance along IS & RDIS & mm & radiograph & 4.0 & 7.8 & 5.9 & 1.0 \\
Spinal cord length along IS from T12 to L4 & SPIS & mm & radiograph & 7.0 & 10.9 & 9.3 & 0.8\\
Weight & WEIG & kg & records & 10.0 & 28.0 & 16.4 & 3.7\\
\bottomrule
\end{tabular}
}	
\end{table}
%

We consider two Organs At Risk (OARs), i.e., organs for which exposure to radiation is known to lead to adverse effects: the liver and the spleen. These OARs are particularly interesting because their shape and position is known to vary substantially per individual, and are very hard to predict~\cite{virgolin2018feasibility}. In general, the liver and spleen are not (clearly) visible in historical radiographs (see Fig.~\ref{fig:radiograph}).
For each patient in the cohort, 3D segmentations of the OARs and of the external body (delimited along IS between T10 and S1) were firstly automatically generated (with ADMIRE research software, 2.3.0, Elekta AB, Stockholm, Sweden), then manually checked and corrected by experienced radiation treatment technologists (with Velocity software, version 3.2.0, Varian Medical Systems, Inc. Palo Alto, CA, US), and finally approved by a pediatric radiation oncologist. 

\subsection{Pipeline for automatic phantom construction} 
The outcome of our pipeline is a CT-based phantom, built using recent patient imaging data. The goal is to resemble the anatomy of the historical patient as closely as possible. 

\begin{figure}[h]
\centering
\includegraphics[width=0.95\linewidth]{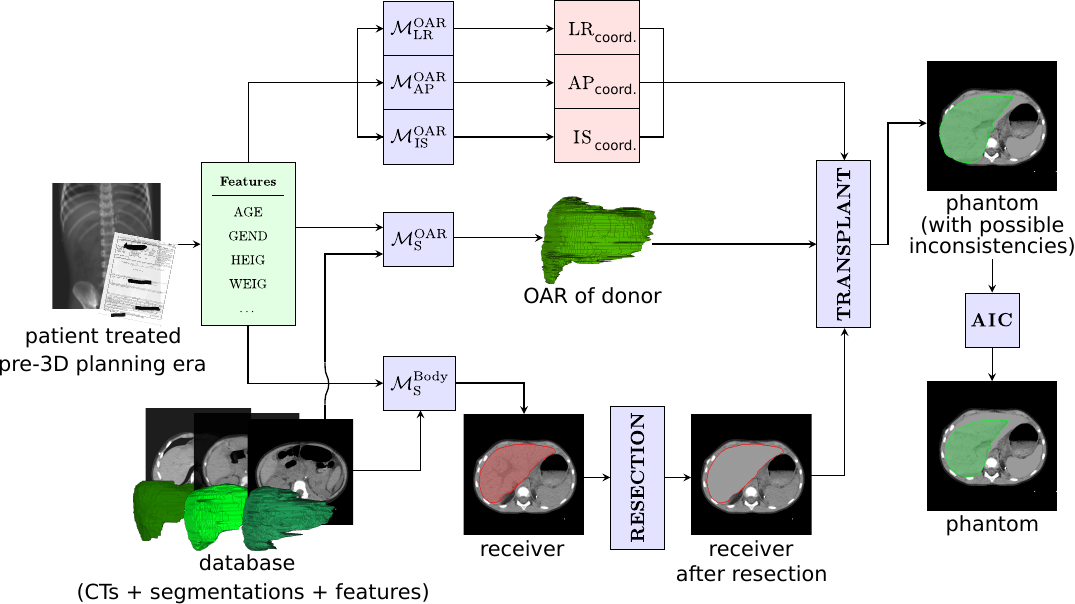}
\caption{Pipeline for automatic phantom building. Pre-trained ML models are used to predict 3D OAR positions as well as what OAR donor segmentation and what receiver CT to retrieve from the database to construct a patient-specific phantom CT (only the liver is considered as OAR in this example). The resected and transplanted OAR is highlighted in red and green, respectively.}\label{fig:pipeline}
\end{figure}

The pipeline is summarized in Fig.~\ref{fig:pipeline}. Given the patient features as input, first an ML model $\mathcal{M}^\text{Body}_\text{S}$ is used to predict which CT is most resembling in terms of overall body shape (for the abdominal region). We call this CT the ``receiver''. Then, the OARs of the receiver are ``resected''. Resection of an OAR is performed by setting the voxels of the receiver that belong to the OAR to Hounsfield Unit (HU) values that represent generic soft abdominal tissues (this is a parameter, we used 78, as done in the phantoms of the University of Florida/National Cancer Institute\cite{geyer2014uf}). Subsequently, for each OAR, its center of mass position is predicted using dedicated models, i.e., one for the LR ($\mathcal{M}^\text{OAR}_\text{LR}$), one for the AP ($\mathcal{M}^\text{OAR}_\text{AP}$), and one for the IS position ($\mathcal{M}^\text{OAR}_\text{IS}$). 
A fifth model ($\mathcal{M}^\text{OAR}_\text{S}$) is used to predict which OAR segmentation to retrieve among the ones available based on a shape-focused metric (described in Sec.~\ref{sec:datasets}). 
We choose to adopt separate models because the features related to one metric may be independent from the ones related to another metric, e.g., we expect features related to the LR direction to be important for the LR coordinate of the center of mass of an OAR, but not (or substantially less) for its AP or IS coordinates and for determining what segmentation has the most promising shape.

We refer to the CT that is chosen by $\mathcal{M}^\text{OAR}_\text{S}$ to provide the OAR segmentation as a ``donor''. Next, the segmentation is ``transplanted'' into the receiver, using the predicted position. Transplantation is achieved similarly to resection: we add the OAR segmentation to the set of segmentations of the receiver, placed in the predicted position, and we set the HU values of the voxels in the receiver that belong to the OAR segmentation to the respective HU values from the donor scan.

A final step consists of correcting the phantom for possible anatomical inconsistencies with an Anatomical Inconsistencies Correction (AIC) procedure. This is because the ML models can in principle predict to place segmentations in positions that result in non-realistic anatomy (e.g, OARs overlapping with each other). An optimizer is therefore used to subsequently resize and re-position the OARs to correct for anatomical inconsistencies, with minimal corrections so as to compromise the quality of the ML model predictions as little as possible (more details are given later, in Sec.~\ref{sec:aic}).


To store CTs and segmentations sets, we relied on the Digital Imaging and Communications in Medicine standard (DICOM). Segmentation sets were stored in DICOM RT-Structure sets. To handle this data format, we used the pydicom package for Python (https://pydicom.github.io/). We do not provide full details on the implementation here (e.g., which fields of the DICOM files are changed and how), but our source code is available at: http://github.com/marcovirgolin/APhA.

\subsection{Machine Learning}\label{sec:ml}
This section describes how the ML models in the pipeline are trained. Firstly, we present how datasets for supervised learning are built. Secondly, we present the ML algorithms that we selected, and their hyper-parameters. Finally, we describe the learning strategy, i.e., how we train, tune, and validate the models.

\subsubsection{Datasets for training}\label{sec:datasets}
To train the ML models we prepare separate datasets to learn each ML model independently. As aforementioned, we learn separate models under the assumption that 3D metrics are (largely) independent from each other.
Hence, $1 + 4\times \#\text{OARs} $ datasets are needed: one to learn how to pick the receiver CT, three for each OAR to predict the OAR position, and one more for each OAR to predict which segmentation to retrieve.



We perform numerical feature normalization by z-scoring~\cite{jain2005score}, i.e., each feature is normalized by subtraction of the mean and division by the standard deviation. Since patient gender is categorical, we set it to a binary value: 0 for females and 1 for males. As no correlation coefficient above 0.9 was found between the features (Fig.~\ref{fig:corrmat}), we do not exclude any feature. Z-scoring is also applied to the target variable $\mathbf{Y}$.

We construct two types of datasets. One type is used for OAR position modeling along three orthogonal directions (i.e., LR, AP, IS). The other type is for donor and receiver retrieval.
The preparation of the datasets for OAR position modeling is straightforward. We define the OAR position relative to a common landmark: the center of the L2 vertebra. In other words, the OAR position in LR, AP, or IS direction is a signed, one-dimensional distance from the center of the L2 vertebra to the center of mass of the OAR's segmentation, measured in mm.

For the datasets concerning the retrieval of a representative OAR segmentation, a measure of segmentation similarity needs to be defined as target variable. Because a measure of similarity is defined between pairs of segmentations, each example needs to be defined over a pair of patients, leading to a total of $\#\text{patients} (\#\text{patients}-1)/2$ training examples. 
For the features, we use a pairwise version of the features defined as the absolute difference of the features of patients $p$ and $q$: $x_{p,q} = | x_p - x_q |$, with $x_p$ being the feature $x$ of patient $p$. For the target variable, several possibilities to address segmentation similarity have been proposed in literature. For example, a possibility is to consider similarity in OAR weight\cite{xie2017computational}. 
We do not rely on OAR weight nor volume because they do not account for similarity of shape.
Another commonly adopted option to assess OAR similarity is the use of the volumetric Dice-S\o rensen Coefficient (DSC)~\cite{dice1945measures,virgolin2018feasibility,zou2004statistical}.
However, the DSC still has limitations, because it is segmentation-volume dependent.
We therefore rely on the recently introduced surface Dice-S\o rensen Coefficient (sDSC)~\cite{nikolov2018deep}, which considers only significant millimetric deviations between the surfaces of the segmentations to evaluate similarity. In particular, the sDSC uses a threshold parameter $\tau$ that expresses what deviations are acceptable (e.g., as part of inter-observer variability). 
To choose $\tau$, we consider that the median CT slice thickness in our database is 2.5mm. Since we deal with inter-patient OAR segmentations, we doubled this value, and adopted $\tau=5.0$mm. 
Because we predict OAR positions separately for each OAR, the sDSC is computed after aligning the segmentations on their center of mass, i.e., to maximally focus on the shape. Furthermore, we consider sDSC values as percentages (0 to 100).


\subsubsection{Loss function for learning}\label{sec:mae}

As loss function to train and validate the ML algorithms we consider the  Mean Absolute Error (MAE), i.e.,

\begin{equation*}
    \text{MAE}( y, \hat{y} ) = \sum_{i=1}^n | y_i - \hat{y}_i |,
\end{equation*}

where $y$ and $\hat{y}$ are $n$-dimensional vectors of ground-truth values and ML predictions respectively. Clearly, the MAE needs to be minimized. We choose to use the MAE because it is relatively simple to interpret, as it preserves the unit of measurement of the metric at hand (mm for OAR positions, \% for sDSC between OAR segmentations), and weighs all errors equally.

For OAR positions, computing the MAE is straightforward both at training and at test time. Given $n$ cases, it suffices to measure the absolute difference between each ground-truth position and its prediction, and take the average upon all examples.

The sDSC is defined between pairs of patient OAR segmentations. Here, ML algorithms predictions correspond to estimations of sDSC between OAR segmentation pairs, and MAE minimization means to correctly predict the sDSC. 
At training time, the MAE between predictions and ground-truth sDSC is measured.
The measurement of validation MAE is more involved. In particular, when validation is needed for a patient who was not included in the training data, i.e., to obtain a segmentation for a test patient, we firstly create pairwise features (Table~\ref{tab:features}) between that patient and each other patient in the database (as absolute feature differences as described in~\ref{sec:datasets}). Subsequently, the model uses the pairwise features as input, and produces predictions of sDSC between the OAR segmentation of the test patient and the one of each other patient (see the input of $\mathcal{M}^\text{OAR}_\text{S}$ and $\mathcal{M}^\text{Body}_\text{S}$ in Fig.~\ref{fig:pipeline}). Now, we do not compute the MAE between these predictions and the ground-truth sDSCs as test error, because the predictions do not correspond to any particular segmentation in the database. Rather, they give an indication of how the OAR segmentation of each patient will fare against the OAR segmentation of the test patient in terms of sDSC. Therefore, we proceed by retrieving the segmentation from the database that has the largest predicted sDSC with the test patient. Finally, we evaluate the actual sDSC between these two segmentations, and report that.

Note that the maximum score of OAR shape similarity achievable is not necessarily 100\%, rather, it is the maximum sDSC that can be obtained with the OAR segmentations available in the database. To frame the future comparisons that involve different metrics, i.e., OAR positions and OAR shape, we define $\varepsilon_\text{sDSC} = 100 - \text{sDSC}$, which is minimal (0\%) when the sDSC is maximal (100\%), and is maximal (100\%) when the sDSC is minimal (0\%). This way, both errors in OAR positioning and in OAR shape retrieval can be considered as a metric to be minimized.

\subsubsection{Algorithms}\label{sec:mlas}
We compare a total of five regression ML algorithms: Least-Angle RegreSsion (LARS) \cite{efron2004least}, LARS using the Least Absolute Shrinkage and Selection Operator (LASSO) \cite{tibshirani1996regression}, Random Forest (RF) \cite{breiman2001random}, traditional Genetic Programming (GP-Trad) \cite{koza1994genetic,poli2008field}, and the Genetic Programming instance of the Gene-pool Optimal Mixing Evolutionary Algorithm (GP-GOMEA) \cite{virgolin2017scalable,virgolin2019model}. These algorithms are interesting because they include hyper-parameters that can be used to prevent overfitting, which is a likely scenario when the data is limited as in many medical applications, including ours.

LARS and LASSO are widely used approaches that work based on the assumption that the features can be combined linearly, and include penalization metrics to avoid overfitting. LASSO also performs feature selection. Both algorithms build models that can be read as mathematical expressions consisting of a linear combination of the features, and are therefore considered interpretable~\cite{guidotti2018survey}. We optimize the hyper-parameter $\lambda$, i.e., the penalization factor, as described in Sec.~\ref{sec:learning-strategy}. We use the implementation of the package glmnet, in R~\cite{team2013r,friedman2009glmnet}.

RF is also widely used, but it is different from the previous two as it allows non-linear modeling. RF builds a model as an ensemble of decision trees to reduce variance in estimation and thus overfitting. Because the model is an ensemble, it is considered not interpretable~\cite{guidotti2018survey}. We optimize how many features are randomly chosen when building the nodes that compose each decision tree (\emph{mtry}), and the minimum number of data samples a node should represent (\emph{min. node size}), the same way we optimize $\lambda$ for LARS and LASSO (see Sec.~\ref{sec:learning-strategy}). We use the R implementation known as ranger~\cite{wright2015ranger}.

The Genetic Programming algorithms are interesting because, like LARS and LASSO, they deliver models in the form of mathematical expressions, but they can include non-linear feature combinations. These algorithms work by loosely mimicking the concept of Darwinian evolution, i.e., by iterative recombination of candidate mathematical expressions made of atomic functions, and selection of the fittest. Recombination in GP-Trad is highly stochastic, whereas the one of GP-GOMEA includes information theory-based mechanisms to estimate what patterns of atomic functions should be preserved during recombination attempts. GP-GOMEA was recently found to work particularly well when small, yet accurate expressions are needed~\cite{virgolin2019model}.
To reduce the number of hyper-parameters, we use these algorithms within a scheme of interleaved runs of increasing capacity (described in sec~\ref{sec:parameters}). We use the \CC{} implementation of GP-Trad and GP-GOMEA (https://github.com/marcovirgolin/GP-GOMEA).

\subsubsection{Hyper-parameter settings}\label{sec:parameters}
Table~\ref{tab:hyperparams} shows the hyper-parameters used by the ML algorithms. LARS and LASSO use $\lambda$ to penalize complex models. 
For RF, we use the default relatively large number of trees (given the datasets at hand) of 500~\cite{probst2017tune}.

For GP-Trad and GP-GOMEA, the function set $\mathcal{F}$ defines which functions to use as model components (tree nodes). The division operator $\div_\text{A}$ is the analytic quotient~\cite{ni2013use}, which not only guarantees that the divider can never be null, but also ensures smoothness (in contrast with the protected division operator~\cite{poli2008field}, which can harm generalization~\cite{ni2013use}).
The logarithm operator is protected to avoid infeasible computations~\cite{koza1994genetic}.
The Ephemeral Random Constants (ERC) are constants for which the value is set by uniform sampling from a defined interval~\cite{poli2008field}. 
Mathematical expressions are encoded as parse trees in GP-Trad and GP-GOMEA. We set a small tree height to keep the resulting mathematical expressions short and readable (we found that larger three heights can result in hard to read expressions), and to prevent overfitting.

The number of candidate expressions to evolve, i.e., the population size, is a sensitive parameter for GP algorithms. We run GP-Trad and GP-GOMEA using the Interleaved Multistart Scheme (IMS), a method that interleaves multiple runs with increasing population size. We set the number of sub-iterations between runs, $g_\text{IMS}$, to 4 as has been reported to work well on benchmark problems~\cite{virgolin2017scalable,virgolin2019model}.
Since the IMS can in principle run forever, we set a time limit of 60 seconds. We found this limit to be reasonable because the datasets are small and evaluations are fast, and because the other ML algorithms take only a few seconds to execute. We also preliminarily observed that increasing the time limit (e.g., to 5 or 10 minutes) does not alter the results in a significant way.
For further details on GP-Trad, GP-GOMEA, the IMS, and other hyper-parameters, the reader is referred to the seminal paper on GP-GOMEA for regression~\cite{virgolin2019model}.

\begin{table}
\caption{Hyper-parameters of the ML algorithms. The subscript ``tune'' means that the hyper-parameter setting is subject to optimization with 5-fold cross-validation grid-search among the listed values.}\label{tab:hyperparams}
\centering
\scalebox{0.9}{
\begin{tabular}{lcc}
\toprule
Algorithm & Hyper-parameter & Settings \\
\midrule
LARS and LASSO 
 & $\lambda_\text{tune}$ & $10^{-10}, 10^{-9}, \dots, 10^{10}$\\
\hline
RF 
 & nr. trees & 500\\
 & min. node size\textsubscript{tune} & $5, 10, \dots, 20, 25$\\
 & mtry\textsubscript{tune} & $1, 2, \dots , \frac{ \text{ \# features } }{2}$\\
\hline
\multirow{2}{*}{\makecell[l]{GP-Trad and \\GP-GOMEA}}
 & $\mathcal{F}$ & $\{ +, -, \times, \div_\text{A}, \exp, \log_\text{P} \}$\\
 & $\text{ERC}$ & $\mathbb{U}[ -10, 10 ]$ \\
 & tree height & 2 \\
 & $ g_\text{IMS} $ & 4 \\
 & time limit & 60s\\
\bottomrule
\end{tabular}
}
\end{table}

\subsubsection{Learning strategy}\label{sec:learning-strategy}
With a limited number of 60 patients available, we perform leave-one-out cross-validation to compute the overall test performance. The leave-one-out cross-validation is performed ``patient-wise'', i.e., all the examples relative to a particular patient $p$ are removed from the training set, and solely used for testing. This is obvious for the datasets on OAR position, as each row corresponds to exactly one patient.
For the datasets on OAR segmentation retrieval, however, each row represents a pair of patients (see Sec.~\ref{sec:datasets}). Therefore, all $\#\text{patients}-1$ (59) rows where $p$ is considered are removed from the training set and put in the test set. This is necessary to avoid a positive bias in the test results.

Within each iteration of leave-one-out cross-validation, for LARS, LASSO, and RF, we perform grid-search hyper-parameter tuning with 5-fold cross-validation upon the training data, to determine the best hyper-parameter values. We use the R package caret for this purpose\cite{kuhn2008caret}. Once the best hyper-parameter settings are found, we train the ML algorithm on the training set using those settings, and test it on the test set. For GP-Trad and GP-GOMEA, we take the best expression found by the interleaved runs started by the IMS. 

Since RF, GP-Trad, and GP-GOMEA are stochastic algorithms, we repeat their execution 10 times, and report the mean result.


\subsection{Anatomical inconsistency correction}\label{sec:aic}
The last step of our pipeline repairs possible anatomical inconsistencies present in the assembled phantoms. We automatically compute possible overlaps between the transplanted OARs (liver and spleen), and between the OARs and the spinal cord segmentation of the receiver, which is available for these patients. To have an additional margin, we enlarge the spinal cord segmentation by 10\% uniformly in all dimensions. Furthermore, we assess if the transplanted OARs stick out of the segmentation of the body of the receiver. If a larger segmentation of the body exists than the common region of interest between T10 and S1 (used to assess body shape similarity to train ML models), that segmentation is used here. Again, for robustness, the body segmentation is shrunk uniformly in all dimensions, by 2.5\%. The values chosen for spinal cord segmentation expansion (10\%) and body segmentation shrinking (2.5\%) were found to deliver pleasing results by visual inspection.

We use a general purpose, derivative-free real-valued optimization algorithm to modify the transplanted OAR segmentations to eliminate the anatomical inconsistencies, with default parameter settings~\cite{bosman2008enhancing,bosman2013benchmarking}. The algorithm is configured to act on both liver and spleen at the same time, with modifications (i.e., optimization variables) that can expand or shrink the segmentations (up to 1.25 and 0.25 of the original volume respectively), as well as reposition their center of mass along LR, AP, IS (up to 10mm for each direction). We set anatomical inconsistencies as hard constraints to satisfy. 
Modifying the OARs to satisfy the constraints deviates from the predictions of the ML models, and can therefore result in less accurate phantoms. Therefore, we use an objective that conflicts with the constraints, in that it attempts to minimize the effect of OAR modifications: for each OAR being corrected, we use the same objective of the ML algorithms to capture the shape of OARs, i.e., the sDSC (again with threshold of 5mm), this time compared to its originally-predicted shape and position. The final objective is given by summing the sDSC of both liver and spleen, to be maximized.

\subsection{Comparing to phantom selection approaches}
As we mentioned in the introduction, it is common practice in phantom-based dose reconstruction to select a phantom from a library according to some criteria. 
As CTs of actual patients can act as phantoms, we  consider several approaches to compare with our pipeline: two methods that simulate state of the art human-designed criteria used to build and select from phantom libraries, random selection, and one method to select a single CT from our database based on ML predictions.

\subsubsection{Human-designed criteria for phantom selection}\label{sec:human-designed-criteria}
The first methods we consider are the criterion used by the University of Texas MD Anderson Cancer Center~\cite{stovall2004genetic,howell2019adaptations}, which we refer to as Human Criterion 1 (HC1), and the criterion used by the University of Florida/National Cancer Institute~\cite{geyer2014uf}, which we refer to as Human Criterion 2 (HC2). We further consider random selection, to see if the other methods are better than random.

The phantoms on which HC1 has previously been used are virtual cuboid shapes with OARs represented as point clouds~\cite{stovall2004genetic,howell2019adaptations}. Only the age of the patient is considered as a feature to manipulate the phantom's representativeness by scaling the cuboids according to guidelines on population data. Furthermore, gender is used to exclude/include gender-specific OARs.
To simulate HC1, i.e., age binning, we cluster our database into age bins, by rounding the age to years. 
This results in 5 bins, with the following distribution: 10 patients of age 2, 18 of age 3, 11 of age 4, 15 of age 5, and 6 of age 6.

For a test patient $p$, we consider the other patients from the database that share the same age bin with $p$. Then, for each metric of interest, we report the average error given by comparing the metric value for $p$ with the one for each other patient that has the same age as $p$. For example, for the assessment of liver segmentation similarity of a patient $p$ that is 3 years old, we compute the sDSC between $p$ and each other patient that is 3 years old, and return the mean.
By doing this, we simulate the fact that an average anatomy has been built using the anatomical information of all patients (different from $p$) that have the same age. We do this because phantoms are built to represent average anatomies. 

The phantom library where HC2 is adopted comprises phantoms made by scaling segmentations acquired from actual patient CT scans, thus they are quite realistic~\cite{geyer2014uf}. For these phantoms, the features considered to build the library were gender, height, and weight (age was not used). HC2 uses these features to select a representative phantom. We simulate HC2 by clustering our database by gender, and by height and weight, using 5 bins for each of the latter two. We use the bin method of the R package binr (http://jabiru.github.io/binr/) for this purpose, with default settings, which results in 5 bins of 12 patients each. Like for HC1, the error for a metric is given by comparing the metric value for the test patient $p$ with the metric values of each patient that shares the same bin of $p$, and taking the average. 

To assess whether there is any merit in using the human-design criteria, we also consider a control method where the OAR positions and segmentations are retrieved uniformly at random from the set of patients, excluding the test patient $p$. Because of the stochastic nature of this approach, each iteration is repeated 10 times, and mean results are computed. In the following, we refer to this method as RAND. 

As we propose a pipeline to build a phantom, it is interesting to assess how it fares against a simpler approach, e.g., to use ML predictions to select a single, overall most representative CT scan. This approach can be related to literature, where new ways to identify which phantom to pick are studied~\cite{virgolin2018feasibility,badouna2012total,stepusin2017assessment}.



While our phantom construction pipeline can predict the different 3D metrics independently (position for each direction, sDSC for each OAR), for a single CT approach, an overall score needs to be defined that expresses how representative a CT scan is. 
The design of such a score is not trivial~\cite{virgolin2018feasibility}. 
For example, a choice needs to be made on whether 5mm along the LR direction is more important or less important than an sDSC loss of 5\%. For the sake of simplicity, we propose a score measure that considers only OAR positions, with equal importance. We choose to focus on OAR position because we recently found it to be the metric that is most correlated with dose accuracy for pediatric abdominal radiation treatment~\cite{wang2019how}.
In detail, we take as best CT the one that is closest to the predictions of the ML models in terms of squared position differences, i.e.,
\begin{equation}\label{eq:sct}
\text{Best CT} = argmin_\text{CT} \sum_{\mathcal{O} \in \text{OARs}} \sum_{\mathcal{D} \in \text{ \{LR,AP,IS\} }} \left( \mathcal{D}^\mathcal{O}_\text{ML} - \mathcal{D}^\mathcal{O}_\text{CT} \right)^2,
\end{equation}
where $\mathcal{D}^\mathcal{O}_\text{ML}$ and $\mathcal{D}^\mathcal{O}_\text{CT}$ are respectively the ML-predicted position and the actually available position in the CT, of the OAR $\mathcal{O}$, along direction $\mathcal{D}$. We denote this approach with sCT. 

Note that sCT is essentially a hybrid between a human-designed criterion, represented by Eq.~\ref{eq:sct}, and the use of ML, of which the predictions are used in the equation.
Lastly, since we found GP-GOMEA to be the overall best performing ML algorithm (shown later in Sec.~\ref{sec:all_ml_criteria}), we used the models found by GP-GOMEA to provide the predictions for sCT.

\subsection{Experimental setup}
We divide experiments into two parts. In the first part, we compare the predictions of the ML algorithms, in terms of MAE. In the second part, we compare the prediction of the overall best performing ML algorithm with the phantom selection approaches, in a similar way. In this second comparison, we include the effect of anatomical inconsistency correction, to assess how much it compromises the accuracy of the predictions of the ML models.

We run the ML algorithms on a machine with two Intel\textsuperscript{\textregistered{}} Xeon\textsuperscript{\textregistered{}} CPU E5-2699 v4 @ 2.20GHz. We assess statistical significance of the results with the Wilcoxon signed-rank test, paired by train-test split (i.e., held-out patient)~\cite{demvsar2006statistical}, and using the Bonferroni correction method to prevent type I errors~\cite{bland1995multiple}. In particular, since we perform pairwise tests between the algorithms for each metric, we assess whether the test $p\text{-value}$ is below a confidence level of 0.05, further reduced by a Bonferroni correction coefficient, to contrast false positive outcomes due to chance.


\section{Results}

\subsection{Comparison of the machine learning algorithms}\label{sec:all_ml_criteria}
The mean (and standard deviation) training and test MAE obtained by the leave-one-out cross-validation of the ML algorithms are reported in Table~\ref{tab:results-mlas}. If multiple results are not found to be statistically significantly worse than the best result, they are considered to be equally good. Note that for consistent use of the MAE, to be minimized, the task of segmentation retrieval uses $\varepsilon_\text{sDSC}$. 

\begin{table}
\caption{Mean training and test MAEs for the ML algorithms on the different OAR-specific regression tasks. Standard deviation is reported in subscript. The MAE for OAR segmentation retrieval is a percentage, the MAE for OAR position estimation is in mm. Results in bold are best in that no other method delivers significantly better ones. The letter ``S'' stands for segmentation retrieval.}
\label{tab:results-mlas}
\centering
\small
\scalebox{0.8}{
\begin{tabular}{cl >{\columncolor[gray]{0.90}}c cccc >{\columncolor[gray]{0.90}}c >{\columncolor[gray]{0.90}}c >{\columncolor[gray]{0.90}}c >{\columncolor[gray]{0.90}}cc }
\toprule
& & \multicolumn{1}{c}{Body} & \multicolumn{4}{c}{Liver} & \multicolumn{4}{c}{Spleen} & \#Best \\
& & S & LR & AP & IS & S & LR & AP & IS & S \\
 \midrule
 \multirow{5}{*}{\begin{sideways}Training\end{sideways}}
& LARS & 
 16.63\tiny{0.08} &
 7.89\tiny{0.24} & 
 4.27\tiny{0.14} & 
 7.10\tiny{0.33} & 
 17.02\tiny{0.07} & 
 \textbf{3.99}\tiny{0.21} & 
 7.21\tiny{0.13} & 
 7.71\tiny{0.19} & 
 15.39\tiny{0.10} & 1\\
 
& LASSO & 
 16.64\tiny{0.08} &
 7.87\tiny{0.36} & 
 4.24\tiny{0.10} & 
 7.28\tiny{0.19} & 
 17.02\tiny{0.07} & 
 4.03\tiny{0.10} & 
 7.24\tiny{0.13} & 
 7.60\tiny{0.13} & 
 15.38\tiny{0.09} & 0\\
 
& RF & 
 \textbf{6.74}\tiny{0.33}&
 8.36\tiny{0.14} & 
 4.87\tiny{0.08} & 
 8.74\tiny{0.09} & 
 \textbf{12.37}\tiny{1.08} & 
 5.44\tiny{0.07} & 
 7.25\tiny{0.14} & 
 9.33\tiny{0.14} & 
 \textbf{7.21}\tiny{0.83} & 3 \\
 
& GP-Trad & 
 19.55\tiny{0.06} &
 \textbf{6.97}\tiny{0.11} & 
 \textbf{3.93}\tiny{0.07} & 
 6.42\tiny{0.10} & 
 15.97\tiny{0.03} & 
 4.01\tiny{0.05} & 
 6.56\tiny{0.13} & 
 7.06\tiny{0.15} & 
 14.08\tiny{0.04} & 2\\
 
& GP-GOMEA & 
 19.55\tiny{0.06} &
 \textbf{6.97}\tiny{0.11} & 
 \textbf{3.93}\tiny{0.07} & 
 \textbf{6.38}\tiny{0.09} & 
 15.96\tiny{0.03} & 
 \textbf{3.95}\tiny{0.05} & 
 \textbf{6.47}\tiny{0.12} & 
 \textbf{7.05}\tiny{0.15} & 
 14.07\tiny{0.04} & 6\\

\hline
\multirow{5}{*}{\begin{sideways}Test\end{sideways}} 
& LARS &
 23.70\tiny{15.31} &
 8.54\tiny{6.83} &
 \textbf{4.82}\tiny{4.57} &
 9.10\tiny{5.33} & 
 \textbf{38.90}\tiny{17.84} & 
 5.18\tiny{3.33} & 
 \textbf{7.42}\tiny{8.06} & 
 8.52\tiny{7.48} & 
 35.12\tiny{14.51} & 3\\
 
& LASSO & 
 22.78\tiny{15.34} &
 8.87\tiny{6.91} &
 \textbf{4.86}\tiny{4.49} & 
 8.98\tiny{5.37} & 
 \textbf{38.98}\tiny{19.11} & 
 5.02\tiny{3.21} & 
 \textbf{7.37}\tiny{8.10} & 
 8.68\tiny{7.47} & 
 36.82\tiny{13.59} & 3\\
 
& RF &
 \textbf{21.38}\tiny{14.59}& 
 8.36\tiny{6.55} & 
 \textbf{4.77}\tiny{4.62} & 
 7.94\tiny{5.73} & 
 41.12\tiny{16.72} & 
 4.98\tiny{3.48} & 
 7.83\tiny{8.00} & 
 8.80\tiny{7.74} & 
 \textbf{33.98}\tiny{15.47} & 3\\
 
& GP-Trad &
 27.08\tiny{16.67} &
 \textbf{7.27}\tiny{6.48} &
 \textbf{4.68}\tiny{4.30} & 
 \textbf{7.23}\tiny{5.89} & 
 \textbf{40.61}\tiny{14.25} & 
 5.23\tiny{3.40} & 
 8.35\tiny{8.22} & 
 8.43\tiny{9.73} & 
 35.70\tiny{10.57} & 4\\

& GP-GOMEA & 
 26.77\tiny{16.75} &
 \textbf{7.26}\tiny{6.48} &
 \textbf{4.78}\tiny{4.36} & 
 7.89\tiny{6.03} & 
 \textbf{39.00}\tiny{19.31} & 
 \textbf{4.56}\tiny{3.49} & 
 \textbf{7.33}\tiny{8.32} & 
 \textbf{8.21}\tiny{9.78} & 
 35.62\tiny{11.43} & 6\\
\bottomrule

\end{tabular}
}

\end{table}

In terms of training performance, GP-GOMEA is overall the best algorithm, as it is not significantly worse than any other for 6 metrics, i.e., OAR position for all directions. RF follows with 3 top performances, in particular for the segmentation retrieval task (S) of all OARs, where it achieves markedly lower errors than all other algorithms. LASSO performs worst, with at least one algorithm significantly outperforming it in each metric.

Regarding the test performance on the patient excluded by the leave-one-out cross-validation, similar to the observed training performance, GP-GOMEA obtains the most significantly best results. GP-GOMEA also generalizes well on predicting which segmentation to select for the liver, but it is inferior to RF when it comes to selecting the segmentations for the body and the spleen. Indeed, for those two segmentations, RF performs better than any other ML algorithm, although the errors at test time are much larger than the ones found at training time. Note that this mismatch between training and test performance can be explained by possible overfitting, as well as by the fact that, at test time, error propagation can happen, since the model prediction is used to retrieve a candidate segmentation from the ones available in the database (this holds for all ML algorithms, see Sec.~\ref{sec:mae}). GP-Trad scores an extra point compared to RF in terms of number of metrics where it is not outperformed significantly. LARS and LASSO generalize quite well at test time, as they are not inferior to the other ML algorithms on (the same) 3 metrics.



\subsection{Machine learning vs. phantom selection approaches}\label{sec:ml+aic_vs_others}
Table~\ref{tab:results-mla-vs-human} shows results comparing the ML algorithm found to perform overall best, GP-GOMEA, to the use of phantom selection approaches. The use of anatomical inconsistency correction upon GP-GOMEA's predictions is also included now (named GP-G\textsubscript{AIC}), because we want the pipeline to construct pseudo-realistic phantoms. We remark that since the use of automatic inconsistency correction does not introduce corrections to the body segmentation, there is no difference between GP-GOMEA and GP-G\textsubscript{AIC} in terms of error for body segmentation retrieval. 

\begin{figure}
    \centering
    \includegraphics[width=0.8\linewidth]{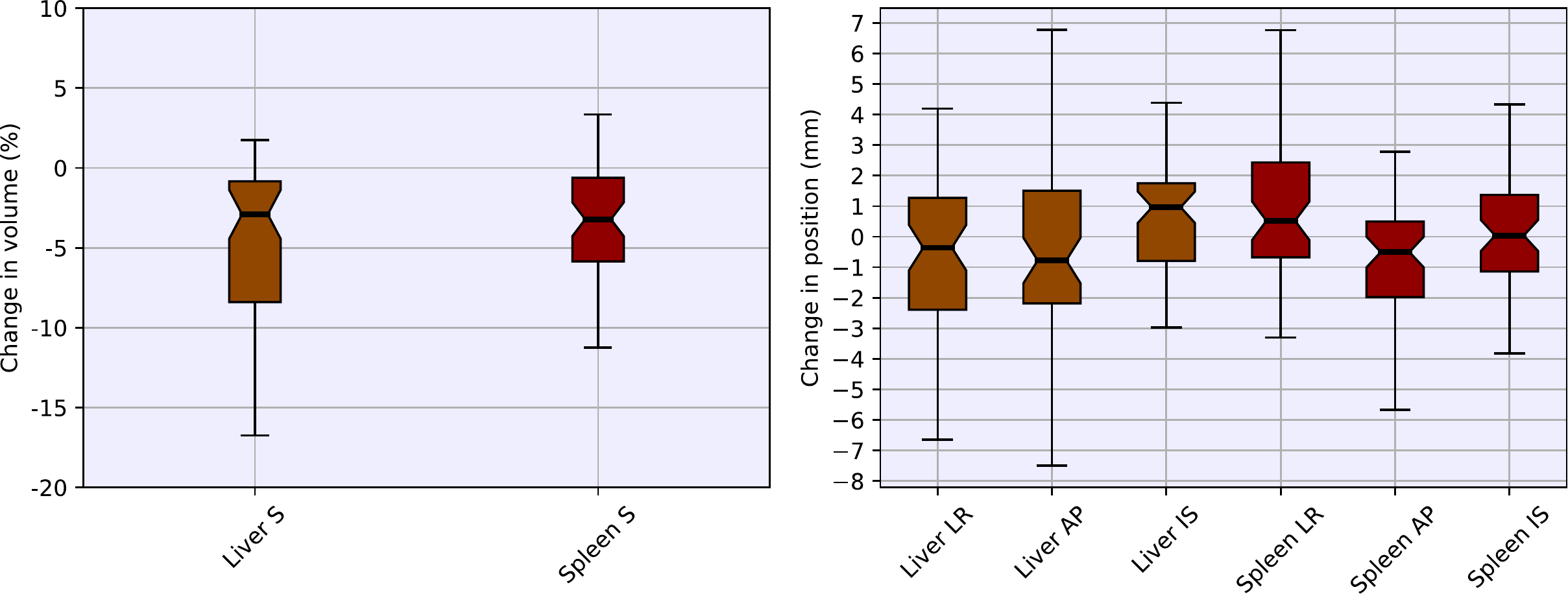}
    \caption{Distribution of the effect of the anatomical inconsistency correction on all phantoms (29/60 are not corrected). OAR shape (S) is corrected by volume enlargement (if $>0\%$) or shrinking (if $<0\%$) uniformly along the three dimensions. Change of center of mass for AP, LR, and IS is in mm. Phantoms where correction is not needed, contribute to OAR shape modification with 0\% (no enlargement nor shrinking), and to OAR position change with 0 mm (no re-positioning). Boxes extend from the 25th to the 75th percentiles, inner bar is the median, and whiskers extend from the 10th to the 90th percentiles.}
    \label{fig:anatomical-correction-extent}
\end{figure}

Out of the nine metrics, the correction-less predictions obtained with GP-GOMEA are generally best, with the only exceptions being the IS position and segmentation retrieval for the spleen (by relatively small errors on average). The use of anatomical inconsistency correction upon GP-GOMEA's predictions, GP-G\textsubscript{AIC}, can change shape and position of the OARs in both considerable (e.g., liver AP and liver S) and minor (e.g., liver LR, spleen IS) magnitude. Figure~\ref{fig:anatomical-correction-extent} shows box plots of the corrections on all phantoms. For our database, the anatomical inconsistency correction was triggered for 31/60 phantoms. The liver is subject to more corrections than the spleen: it is typically shrunk more than the spleen, and its position in AP is subject to large variations. This is not surprising, because the liver is a considerably larger organ than the spleen, and it is more likely to violate the anatomical consistency constraints we imposed. 

Overall, correcting for inconsistencies typically comes at the cost of worsening the accuracy of the ML predictions. GP-G\textsubscript{AIC} has significantly best performance only on 3 metrics (body S, spleen AP, and spleen S). 
For spleen AP and spleen S, performing inconsistency correction leads to better test results compared to not applying corrections. However, the correction algorithm solely optimizes for resolving the inconsistencies while attempting to retain maximum prediction fidelity. Moreover, because sCT obtains also good results on spleen AP and spleen S, it can be argued that these metrics are not modeled sufficiently well by GP-GOMEA to begin with, and thus are more likely to be improved upon by chance.

Although the use of anatomical inconsistency correction after ML prediction typically leads to larger errors, it remains a valuable approach compared to the other methods. GP-G\textsubscript{AIC} is overall best when the correction-less GP-GOMEA predictions are excluded from the comparison. The human-designed criteria HC1 and HC2 perform overall worse than GP-G\textsubscript{AIC}. Despite its simplicity, HC1 performs well for AP and S of the liver. This may be because metrics related to the liver are harder to model by the ML algorithms, and because the use of anatomical inconsistency correction compromises GP-GOMEA's prediction (particularly notable for liver AP). Despite the fact that HC2 is a somewhat more involved criterion compared to HC1 (HC2 considers gender, height, and weight, while HC1 considers only gender and age), it is never found to be competitive on any metric. This could mean that weight and height are not good features when accounting for similarity of internal anatomy in children. This result is in agreement with previous work~\cite{wang2019how} where lack of correlations between dose reconstruction outcomes were found w.r.t. height and weight.

\begin{figure}
\centering
\small
\setlength{\tabcolsep}{0.3mm}
\def\arraystretch{1}
\begin{tabular}{cc}
 & Body S\\
 \begin{sideways}\makebox[0.30\linewidth][c]{\small Probability}\end{sideways} & 
\includegraphics[width=0.5\linewidth]{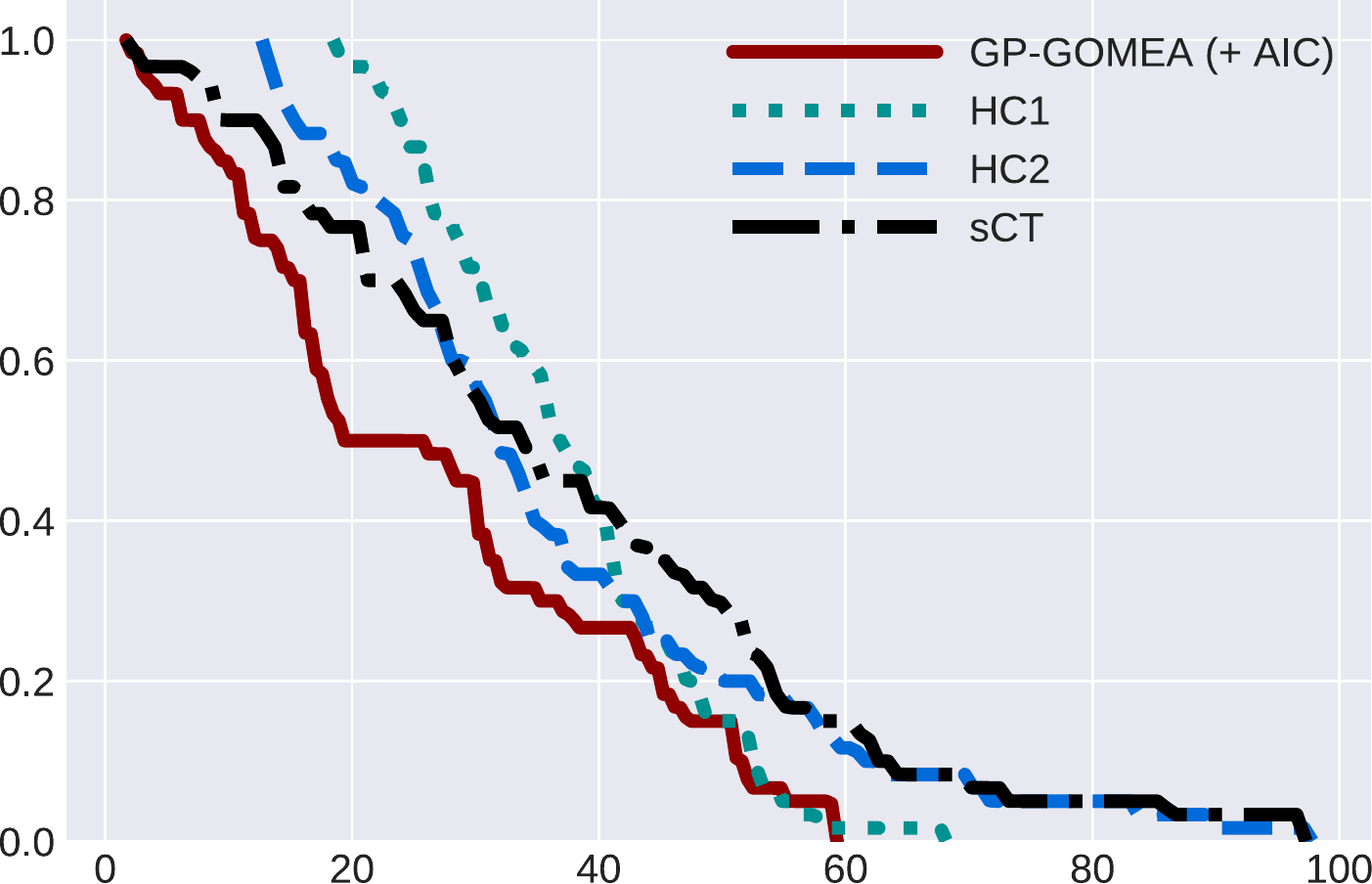}
\\
& {Test MAE}
\end{tabular}
\caption{Reliability functions for prediction of segmentation retrieval (S) of the external body segmentations. The $y$ value is the probability of committing an error equal or greater than the $x$ value. The test MAE is in $\varepsilon_\text{sDSC}$. Note that the anatomical inconsistency correction does not alter the body segmentation. GP-GOMEA leads to much better performance compared to the other approaches.}\label{fig:survplot-body}
\end{figure}

Importantly, in some cases, HC1 and HC2 perform particularly bad. HC1 is particularly inaccurate for spleen S compared to the other approaches, and HC2 leads to notably large errors for spleen IS. In the latter case, HC2 is not found to be better than RAND. Furthermore, both criteria perform poorly on body S (as well as sCT). Figure~\ref{fig:survplot-body} shows the reliability function of the considered approaches for body S. This function shows the likelihood of committing errors of a certain magnitude (for segmentation retrieval, in percentage). A curve is better than the others if it is more on the left and if it decreases more rapidly, since this means that the probability of any error magnitude is lower than the ones of the other methods. The figure illustrates that the model learned by GP-GOMEA achieves this behavior. For example, the probability of a GP-GOMEA's model to predict a body segmentation that has an $\varepsilon_\text{sDSC} > 20\%$ ($\text{sDSC} < 80\%)$ with the actual body of the patient is $60\%$. For HC2 and sCT, that magnitude of error happens more frequently, i.e., in almost $80\%$ of the cases. HC1 performs worse, since errors above $20\%$ are almost certain. On the other hand, rarely, (probabilities around 5\%), HC2 and sCT can retrieve body segmentations that have errors above 80\%.

Regarding sCT, the results indicate that this approach is generally worse than GP-G\textsubscript{AIC}, but better than the human-designed criteria to predict metrics related to the spleen. sCT performs particularly bad with respect to metrics regarding the liver (note in particular liver LR). This is an interesting result because the CT selected by sCT weighs OAR positions equally for liver and spleen (see Eq.~\ref{eq:sct}). We remark that sCT uses the ML models (found by GP-GOMEA) to predict which single CT scan (and accompanying segmentations) to select by means of a hand-designed metric. Essentially, the results confirm our hypothesis that designing a good score to select one CT is not trivial, and that there is added value in constructing a new anatomy by assembling different components into one.

\begin{table}
\caption{Mean test MAEs of the overall best performing ML algorithm GP-GOMEA, also including anatomical inconsistency correction (GP-G\textsubscript{AIC}), and of the phantom selection approaches. Standard deviation is reported in subscript. The MAE for OAR segmentation retrieval is a percentage, the MAE for OAR position estimation is in mm. Results in bold are best in that no other method delivers significantly better ones. Results underlined are best if GP-GOMEA is excluded from the comparison. The letter ``S'' stands for segmentation retrieval.}
\label{tab:results-mla-vs-human}
\centering
\small
\scalebox{0.8}{
\begin{tabular}{cl >{\columncolor[gray]{0.90}}c cccc >{\columncolor[gray]{0.90}}c >{\columncolor[gray]{0.90}}c >{\columncolor[gray]{0.90}}c >{\columncolor[gray]{0.90}}cc }
\toprule
& & \multicolumn{1}{c}{Body} & \multicolumn{4}{c}{Liver} & \multicolumn{4}{c}{Spleen} &  \#Best  \\
& & S & LR & AP & IS & S & LR & AP & IS & S & (Underlined)\\
 \midrule

\multirow{6}{*}{\begin{sideways}Test\end{sideways}} 

& GP-GOMEA & 
 \textbf{26.77}\tiny{16.75} &
 \textbf{7.26}\tiny{6.48} &
 \textbf{4.78}\tiny{4.36} & 
 \textbf{7.89}\tiny{6.03} & 
 \textbf{39.00}\tiny{19.31} & 
 \textbf{4.56}\tiny{3.49} & 
 \textbf{7.33}\tiny{8.32} & 
 8.21\tiny{9.78} & 
 35.62\tiny{11.43} & 7 (--)\\

& GP-G\textsubscript{AIC} & 
 \underline{\textbf{26.77}}\tiny{16.75} &
 \underline{7.91}\tiny{6.27} &
 6.21\tiny{5.42} & 
 \underline{\textbf{8.07}}\tiny{6.06} & 
 \underline{42.20}\tiny{20.29} & 
 \underline{5.18}\tiny{4.04} & 
 \underline{\textbf{7.35}}\tiny{8.30} & 
 8.54\tiny{9.67} & 
 \underline{\textbf{34.10}}\tiny{16.97} & 4 (7) \\

& HC1 & 
 37.54\tiny{10.82} &
 8.88\tiny{6.90} & 
 \underline{\textbf{4.83}}\tiny{4.70} & 
 9.10\tiny{6.13} & 
 \underline{\textbf{43.19}}\tiny{8.35} & 
 5.65\tiny{3.68} & 
 7.96\tiny{7.75} & 
 9.78\tiny{8.20} & 
 37.58\tiny{8.53} & 2 (2) \\
 
& HC2 & 
 36.83\tiny{18.97} &
 9.84\tiny{6.26} & 
 5.18\tiny{4.74} & 
 9.20\tiny{6.48} & 
 49.67\tiny{8.2} & 
 5.54\tiny{4.21} & 
 8.02\tiny{7.69} & 
 13.91\tiny{11.09} & 
 34.76\tiny{10.37} & 0 (0)\\
 
& RAND & 
 45.10\tiny{13.14} &
 11.57\tiny{6.19} & 
 7.11\tiny{4.01} & 
 12.38\tiny{4.48} & 
 44.01\tiny{8.96} & 
 7.70\tiny{3.29} & 
 11.14\tiny{7.22} & 
 13.52\tiny{7.05} & 
 35.89\tiny{8.11} & 0 (0)\\
 
& sCT & 
 37.13\tiny{22.00} &
 15.29\tiny{10.02} & 
 7.44\tiny{5.91} & 
 11.45\tiny{9.2} & 
 \underline{42.72}\tiny{18.30} & 
 \underline{4.85}\tiny{3.43} & 
 \underline{\textbf{7.40}}\tiny{8.32} & 
 \underline{\textbf{7.84}}\tiny{9.37} & 
 \underline{\textbf{32.08}}\tiny{12.28} & 3 (5)\\
 
\bottomrule

\end{tabular}
}

\end{table}

\subsection{Model interpretability}
As we mentioned before, the added value of utilizing the linear ML algorithms (LARS and LASSO), and the GP algorithms (GP-Trad and GP-GOMEA) is that their models are mathematical expressions which, if simple enough, can be interpreted. Interpretability can play a crucial role in determining whether ML can be applied in some clinical settings.
For RF, the interpretation of the model is essentially impossible, as it returns an ensemble of (500) decision trees. Similarly, this would not be possible with other popular techniques like deep learning~\cite{lecun2015deep}, and boosting algorithms~\cite{friedman2001greedy}. 

We report all the best-at-test-time models found by GP-GOMEA at http://bit.ly/2Za4ESy. Here, we report two of those models, that are remarkably simple. For the prediction of which body segmentation to retrieve, the model found most frequently in 10 repetitions is (assuming the target variable and the features are normalized):

\begin{equation*}
0.420 \times \left( \textbf{ADAP}+\textbf{ADLR} + \textbf{SCIS} \right).
\end{equation*}

Notably, this model is just a scaled sum of the abdominal diameters (ADAP and ADLR) and of the spinal cord length (SCIS). Essentially, the model uses an equal contribution (features are normalized) of features capturing information of size relative to the three dimensions LR, AP, and IS, to predict which body segmentation to retrieve. This seems a reliable, simple, and reasonable method to select a representative body segmentation.

A second model we showcase here is the one for the prediction of what spleen segmentation to retrieve:

\begin{equation*}
2.718^{ \textbf{AGE} } \times 0.057 \times \textbf{SCIS}.
\end{equation*}

It is interesting to see that spleen shape is found to be related to age by an exponentiation. This can be considered reasonable for our cohort, because the age of our patients is between 2 to 6 years, where anatomical development is rapid. The length of the spinal cord in IS further weighs the prediction. This feature seems also reasonable to consider, because it captures information relative to the size of the abdomen in IS, and because the spleen is located nearby the spinal cord. Albeit understandable, reasonable, and well-performing, it is arguably unlikely for humans to invent models like these.

\section{Examples of automatically constructed phantoms}
Following the quantitative results presented in the previous sections, we present some qualitative ones:
examples of constructed phantoms. These qualitative results complement the quantitative ones, as the positioning of OARs and their shapes can be visually evaluated in the context of the receiver CT.

\begin{figure}
\centering
\setlength{\tabcolsep}{0.3mm}
\def\arraystretch{0.3}
\begin{tabular}{m{0.02\linewidth}m{0.22\linewidth}m{0.22\linewidth}m{0.22\linewidth}m{0.22\linewidth}}
 & \begin{center}Axial\end{center} 
 & \begin{center}Coronal\end{center}
 & \begin{center}Sagittal\end{center}
 & \begin{center}3D\end{center}\\
 [-3mm]
 
\begin{sideways}Phantom 1\end{sideways}
& \includegraphics[width=3.75cm,height=3.75cm]{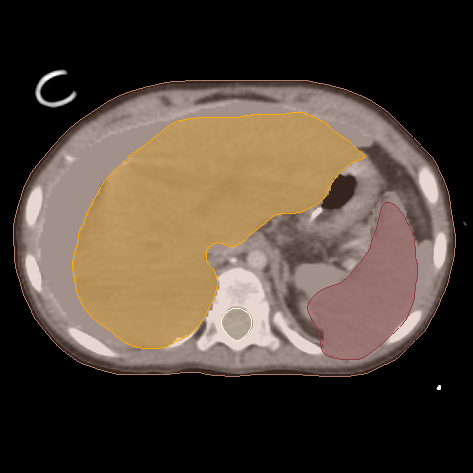} 
& \includegraphics[width=3.75cm,height=3.75cm]{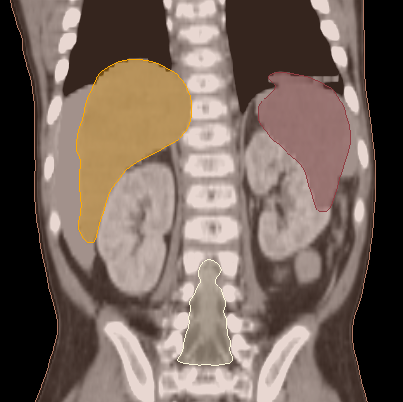} 
& \includegraphics[width=3.75cm,height=3.75cm]{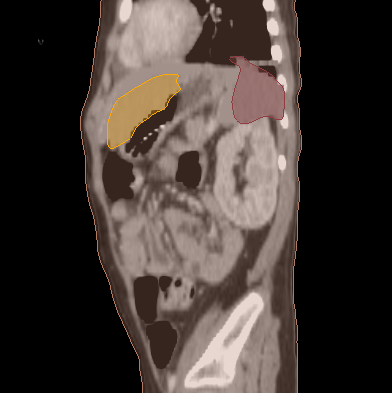} 
& \includegraphics[width=3.75cm,height=3.75cm]{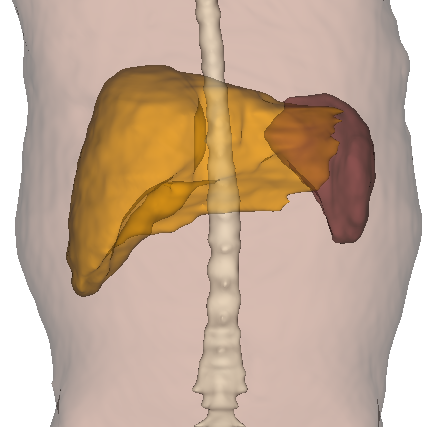}  \\

\begin{sideways}Phantom 2\end{sideways}
& \includegraphics[width=3.75cm,height=3.75cm]{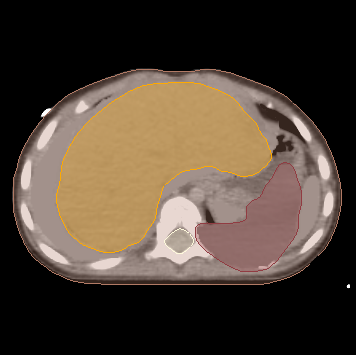} 
& \includegraphics[width=3.75cm,height=3.75cm]{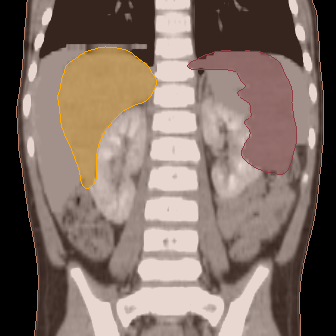} 
& \includegraphics[width=3.75cm,height=3.75cm]{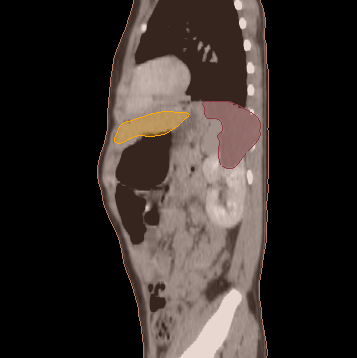} 
& \includegraphics[width=3.75cm,height=3.75cm]{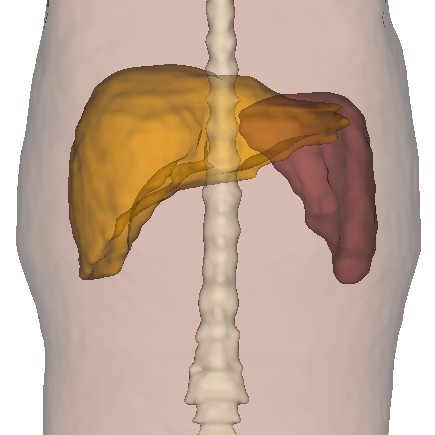}  \\

\begin{sideways}Phantom 3\end{sideways}
& \includegraphics[width=3.75cm,height=3.75cm]{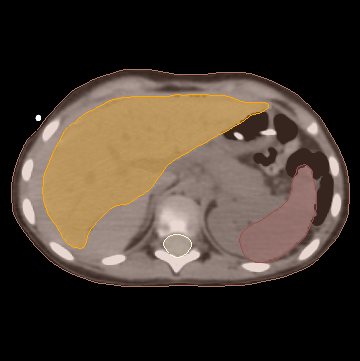} 
& \includegraphics[width=3.75cm,height=3.75cm]{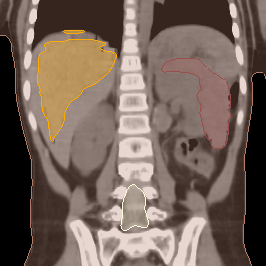} 
& \includegraphics[width=3.75cm,height=3.75cm]{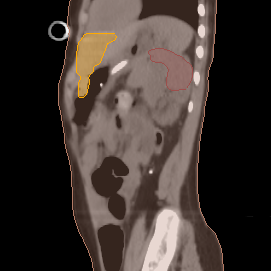} 
& \includegraphics[width=3.75cm,height=3.75cm]{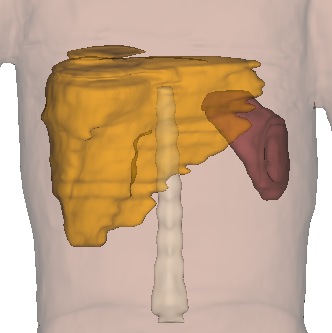}  \\

\begin{sideways}Phantom 4\end{sideways}
& \includegraphics[width=3.75cm,height=3.75cm]{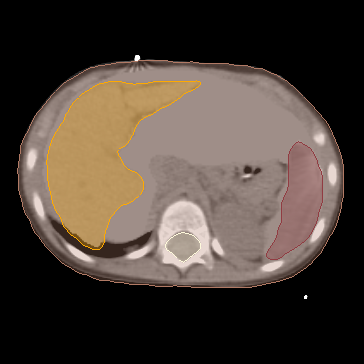} 
& \includegraphics[width=3.75cm,height=3.75cm]{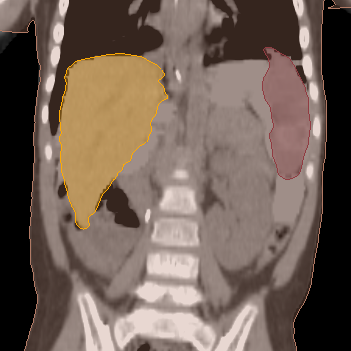} 
& \includegraphics[width=3.75cm,height=3.75cm]{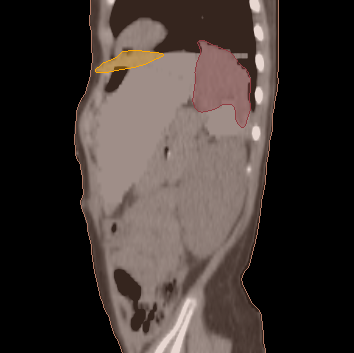} 
& \includegraphics[width=3.75cm,height=3.75cm]{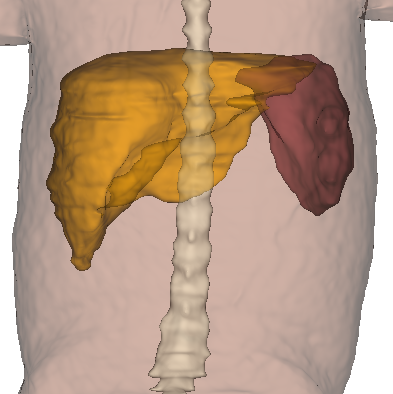}  \\

\begin{sideways}Phantom 5\end{sideways}
& \includegraphics[width=3.75cm,height=3.75cm]{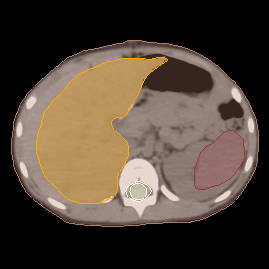} 
& \includegraphics[width=3.75cm,height=3.75cm]{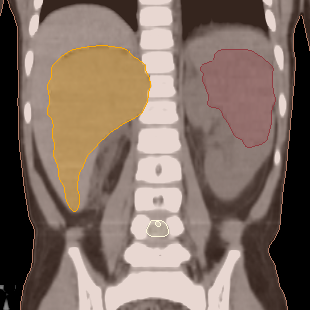} 
& \includegraphics[width=3.75cm,height=3.75cm]{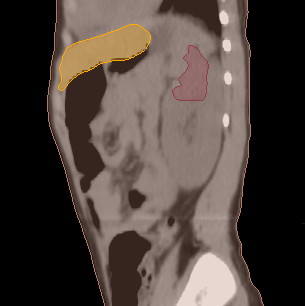} 
& \includegraphics[width=3.75cm,height=3.75cm]{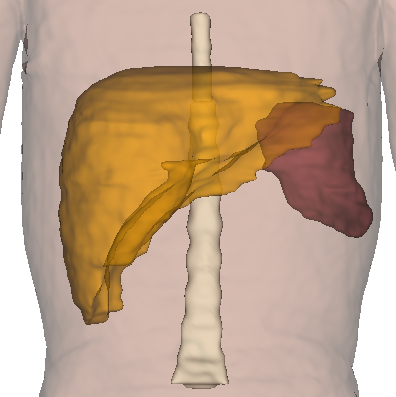}  \\

\\

\end{tabular}
\caption{Examples of phantoms constructed with our pipeline. CT snapshots are chosen to attempt to display both liver (in ocher) and spleen (in crimson red). Axial views are in IS, coronal views and 3D views are in AP, sagittal views are in LR.}\label{fig:example-phantoms}
\end{figure}

Figure~\ref{fig:example-phantoms} shows 5 phantoms generated with our pipeline, where GP-GOMEA was used to train all models. The images are created by using 3D Slicer~\cite{fedorov20123d} with module SlicerRT~\cite{pinter2012slicerrt}. These phantoms have been generated for a test patient that was excluded from the ML training process. The predicted liver and spleen segmentations are highlighted in the receiver CT. Note that the segmentations of the external body and of the spinal cord shown in the figure can extend beyond the commonly available region of interest used for training (from T10 to S1). The receiver CT's original liver and spleen can also be identified (in part), as their not-overridden voxels are uniformly set to a specific value (78 HUs as done for the phantoms of the University of Florida/National Cancer Institute~\cite{geyer2014uf}) in the resection step. 
Overlaps of the transplanted OAR can still happen with respect to other organs which are not considered in the training and correction process (e.g., spleen overlapping with the kidneys). Note also the presence of some further remaining limited anatomical inconsistencies, which are not detected by our algorithm (e.g., small overlaps with bony anatomy). 

Our current Python implementation takes about 5 minutes to generate a phantom if no anatomical inconsistencies are found, with resection and transplant taking the most time (we expect that parallelizing voxels' HU value overwrites will markedly reduce running time). If anatomical inconsistency correction needs to be performed, the whole pipeline can take from a few minutes to a few hours, depending on how complex the correction is for the optimizer, and on the hardware (the optimizer is highly parallelizable). As mentioned in Section~\ref{sec:ml+aic_vs_others}, for our database the anatomical inconsistency correction triggered on half the phantoms. However, in our opinion only roughly half of those cases (so 1/4 of the total number of phantoms) had really noticeable anatomical inconsistencies, e.g., large OAR overlaps, or OARs exceeding the body boundaries, that required corrections of large magnitude. In the other cases, inconsistencies were more subtle (e.g., small OAR overlap), and caused corrections of small magnitude.

\begin{figure}
    \centering
    \setlength{\tabcolsep}{0.3mm}
    \def\arraystretch{0.3}
    \begin{tabular}{m{0.02\linewidth}m{0.3\linewidth}>{\centering\arraybackslash}m{0.3\linewidth}}
    \\
    & \begin{center}Axial\end{center} 
    & \begin{center}3D\end{center}\\
    [-3mm]
    \begin{sideways}
    Liver correction
    \end{sideways} & 
    \begin{tabular}{l} \includegraphics[width=1.0\linewidth,height=1.0\linewidth]{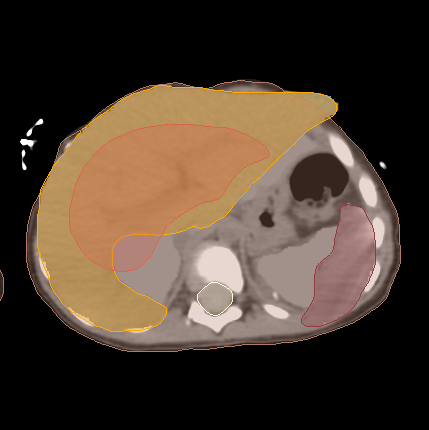} \end{tabular} &
    \begin{tabular}{l} \includegraphics[width=1.0\linewidth,height=1.0\linewidth]{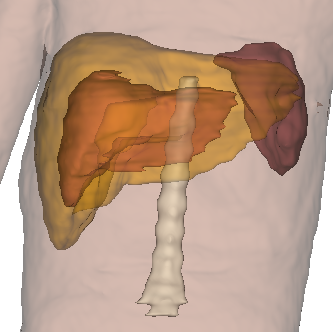} \end{tabular}
    \\
    \begin{sideways}
    Spleen correction
    \end{sideways} &
    \begin{tabular}{l} \scalebox{-1}[1]{\includegraphics[width=1.0\linewidth,height=1.0\linewidth]{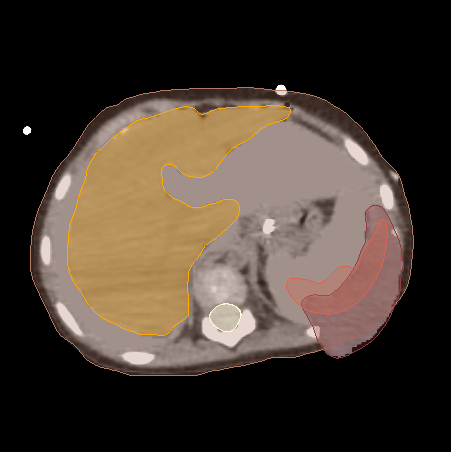}} \end{tabular} &
    \begin{tabular}{l} \includegraphics[width=1.0\linewidth,height=1.0\linewidth]{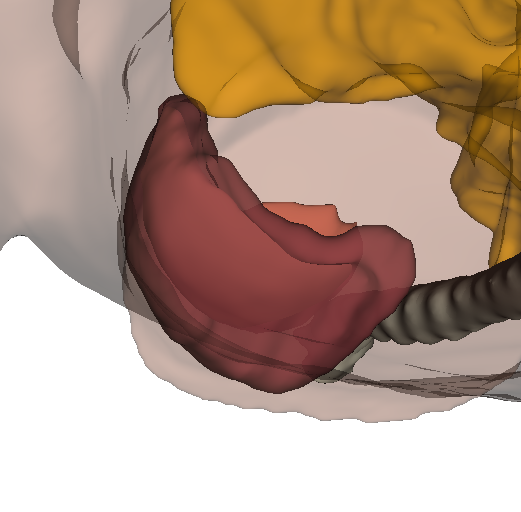} \end{tabular}
    \\
    \end{tabular}
    \vspace{0.7mm}
    \caption{Examples of anatomical inconsistency correction tackling OARs being partly positioned outside of the body segmentation. Pre-correction liver is in ocher, post-correction in orange; pre-correction spleen is in crimson red, post-correction is in salmon pink. in  Top: Axial (left) and 3D (right) view of large anatomical inconsistency involving the liver. Bottom: Axial (left) and 3D (right) view of large anatomical inconsistency involving the spleen. Note that this axial view is different from the ones in other figures as it is taken in superior-inferior direction (spleen displayed on the left), to be consistent with the 3D visualization. The 3D view shows the back of the patient from head to toes.}
    \label{fig:aic_in_action}
\end{figure}

Figure~\ref{fig:aic_in_action} shows two examples of anatomical inconsistency correction for cases where inconsistency can be considered large, displaying phantoms pre- and post anatomical inconsistency correction, one to correct the liver, and one to correct the spleen.


\section{Discussion}
We have presented a new take on phantom construction: a fully-automatic ML-based pipeline that assembles a patient-specific phantom using a database of delineated 3D CT scans, given the features of a patient. We performed experiments upon data of 60 pediatric patients including imaging of the abdomen, and focused on tailoring the position and shape of the liver and the spleen. Our experimental results strongly suggest that our approach leads to much more representative phantoms than using established human-designed criteria, and than using ML to predict a single best CT scan (according to a reasonable notion of overall anatomical similarity). We compared several ML algorithms to provide accurate models for the pipeline, and found GP-GOMEA to deliver overall best performance and models that can also be interpreted, which may be helpful for researchers and clinicians alike to trust their use in a clinical setting.


One clear limitation of our work is that we could only employ a small database of 60 patients. It can be reasonably expected that, by increasing the database size, both the errors of the ML models, and the onset of large anatomical inconsistencies, will be reduced. To this end, we are currently working on expanding the number of institutes contributing to the database. Furthermore, we are working on extending the number of OARs to build the phantom (e.g., heart, kidneys) and on extending the region of interest (e.g., abdomen and thorax), for a cohort including older children (2 to 8 years). 

\begin{figure}
    \centering
    \setlength{\tabcolsep}{0.3mm}
    \def\arraystretch{0.3}
    \begin{tabular}{cc}
    \includegraphics[width=7cm,height=7cm]{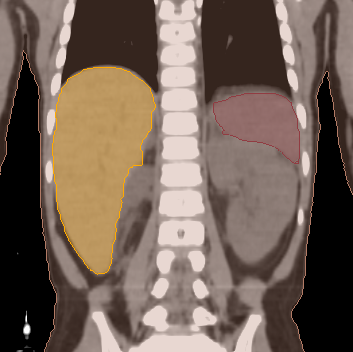}
    &
    \includegraphics[width=7cm,height=7cm]{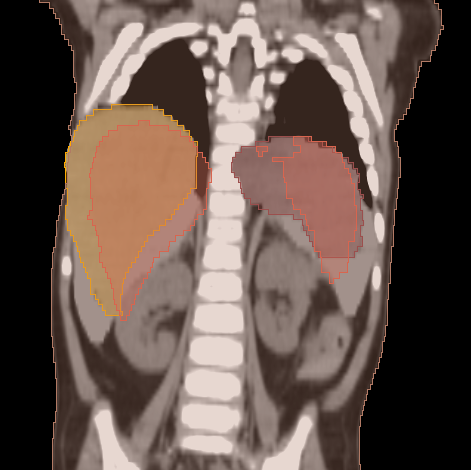}
    \end{tabular}
    \caption{Example of the limitations of anatomical inconsistency correction when applied to a coarse prediction of the body segmentation. Colors as in Fig.~\ref{fig:aic_in_action}. Left: Actual anatomy of the patient. Right: Proposed phantom, including corrections. The receiver CT is quite smaller than the actual CT in IS dimension. The anatomical inconsistency correction shrunk and relocated the liver that was exceeding the body boundaries, and moved the spleen away from the spinal cord. However, liver and spleen are placed too high with respect to the underlying anatomy for it to be realistic.}
    \label{fig:aic_not_perfect}
\end{figure}

The automatic anatomical inconsistency correction method may still warrant further improvement. Although we could always resolve the constraint, often without excessively compromising the predictions of the ML models, anatomical inconsistencies can still be present. This is because the constraints are not comprehensive enough. For example, Fig.~\ref{fig:aic_not_perfect} shows a phantom for which our correction does not violate any of the specified constraints (liver and spleen do not overlap with each other nor with the spinal cord, and the organs do not exceed the body contour) yet the anatomy remains unrealistic because the organs are placed too high with respect to the receiver CT. In fact, this is because it is the prediction of the receiver CT that does not work well for this patient, as it retrieves a body which is quite shorter (14 cm smaller SCIS, sDSC of $60\%$) compared to the actual body of the patient. Fortunately, this is the only phantom out of 60 where such an evident inconsistency was found and likely the probability of this happening only reduces with a growing database size. Still, we intend to study how to further improve our correction method by attempting to craft more constraints, as well as by including more OARs (e.g., the lungs), which then will need to not overlap with each other.

We remark that, in general, phantoms do not need to be anatomically realistic for the purpose of dose reconstruction, especially when doing large dose reconstructions of populations. E.g., as aforementioned, the University of Texas/MD Anderson Cancer Center uses virtual cuboid phantoms uniformly made of water-equivalent material, where internal organs are represented by point clouds~\cite{stovall2004genetic,howell2019adaptations}. However, realistic surrogate anatomies have arguably a larger chance of being considered familiar and trustworthy by clinicians, whenever e.g., an expert opinion is needed to go beyond simple statistical measurements, and really visualize how the radiation dose distribution may impact healthy tissues.

Another limitation of this study is that we used HC1 and HC2 to simulate the process of phantom selection in state of the art phantom libraries on our own database of CT scans and organ segmentations, rather than using phantom libraries. Indeed, the phantoms in those libraries are built to follow statistics relative to large populations. We did investigate the use of the library of the University of Florida/National Cancer Institute, which is of public access, instead of our database, for HC2. We however found that selecting those phantoms leads to no better results than using our database (the body segmentation was not considered because a region of interest between T10 and S1 is not readily available for those phantoms). In particular, it was found to be significantly better for half of the 3D metrics (LR and S for the liver, LR and IS for the spleen), but worse for the other half.  We also found the use of the actual library to be always significantly worse than using GP-GOMEA, with exception for LR of the spleen, where it was equivalent.
This may be because the statistics on which those phantoms are based, which come from the United States, are not accurately representing the patients of our cohort, who are Dutch.

In future work we will consider porting our approach to different types of regions of interest (e.g., head for brain tumors) and cohorts (e.g., older patients). Ultimately, we are interested in generating an entire body anatomy for any patient. We believe our approach is very promising because once appropriate features are defined, no modifications in how to train ML algorithms, nor in how the pipeline works, are needed to obtain new phantoms.
Moreover, since the availability of different OAR segmentations is currently limited to what is available in the database, and since delineating new OARs requires specialization, experience, and time, ways to use ML to deform an existing OAR template into a patient-specific segmentation could be worth investigating~\cite{ng2010reconstruction}.

Finally, for the aim of obtaining 3D dose distributions to relate to the onset of adverse effects, it will be important to validate our pipeline in terms of dose reconstruction accuracy, i.e., by first crafting a highly individualized phantom, and then simulating the radiation treatment plan using such a phantom. This could be performed similarly to our validation analysis, i.e., with cross-validation on recent patients. Dose metrics should be computed on the actual CT, and then compared with the dose metrics computed on the phantom~\cite{virgolin2018feasibility}.

\section{Conclusion}
We have presented a new take on phantom construction that leverages machine learning to assemble existing 3D patient imaging into a new anatomy. Contrary to existing approaches, the pipeline we propose requires no manual intervention except for the initial effort of assembling a database of 3D patient imaging (CTs, segmentations, and patient features), and the measurement of few features of the historical patients on their radiographs. With our approach the problem of finding a globally good metric to represent anatomical categorization, typically faced by phantom libraries, is shifted to train machine learning models for parts of phantoms based on specific 3D metrics. Our experimental results on a database of 60 pediatric cancer patients, focused on liver and spleen, showed that this approach can lead to significantly better anatomical resemblance compared to the use of phantom building criteria that are currently common practice. Positive results were still found after correcting the phantoms for possible anatomical inconsistencies through optimization. Regarding the machine learning algorithm used in the pipeline, we found that GP-GOMEA, a state of the art genetic programming approach, can deliver models that are both accurate and readable. This aspect can be of added value as such models increase the chances of clinicians understanding them better and trusting their use.


%

\subsection*{Disclosures}
Tanja Alderliesten and Peter A.N. Bosman are involved in projects supported by Elekta (Elekta AB, Stockholm, Sweden). Elekta had no involvement in the study design, data collection, analysis and interpretation, and writing of the article.

\acknowledgments 
The authors acknowledge the Kinderen Kankervrij foundation for financial support (project \#187), and the Maurits and Anna de Kock foundation for financing a high-performance computing system. We thank dr. Brian V. Balgobind, dr. Irma W.E.M. van Dijk, and dr. Jan Wiersma from the department of radiation oncology of Amsterdam UMC, location AMC, Amsterdam, the Netherlands, and dr. Geert O.R. Janssens and dr. Petra Kroon from the department of radiation oncology of UMC Utrecht Cancer Center, Utrecht, the Netherlands, for providing help in the collection and/or in the assessment of the imaging data used in this work. The authors are grateful to Elekta for providing ADMIRE research software for automatic organ segmentation. We further acknowledge dr. Choonsik Lee from the National Cancer Institute, Division of Cancer Epidemiology \& Genetics, Rockville, Maryland, U.S.A., for details on the phantom library of the University of Florida/National Cancer Institute. This article is based upon, and extends (by introducing the anatomical inconsistency correction method and related results), an SPIE proceedings paper of which an abstract has been recently submitted for consideration to the SPIE conference on Medical Imaging (2020), titled "\emph{Machine Learning for Automatic Construction of Pediatric Abdominal Phantoms for Radiation Dose Reconstruction}", authored by the same authors of this article.


\bibliography{report}   
\bibliographystyle{spiejour}   


\vspace{2ex}\noindent\textbf{Marco Virgolin} is a PhD candidate at Centrum Wiskunde \& Informatica in Amsterdam, the Netherlands, and is enrolled at the Delft University of Technology, Delft, the Netherlands. He holds an M.Sc. in Computer Engineering from the University of Trieste, Italy. Marco is mostly interested in evolutionary and explainable machine learning, with a special focus on genetic programming and symbolic regression. For more information, visit http://marcovirgolin.github.io.

\vspace{2ex}\noindent\textbf{Ziyuan Wang} is a PhD student working in the Radiation Oncology department, Amsterdam UMC, University of Amsterdam. Her research interests include medical imaging analysis and physics of radiation treatment. She has a background in Applied Physics and is currently working on the project ``3D dose reconstruction for children with long-term follow-up: Toward improved decision making in radiation treatment for children with cancer''.

\vspace{2ex}\noindent\textbf{Tanja Alderliesten} received her PhD in computer science in 2004 from the Utrecht University in the Netherlands. Currently, she is a tenured senior researcher at the Department of Radiation Oncology of the Amsterdam UMC, University of Amsterdam, the Netherlands. The focus of her research is translational in nature and primarly concerns the development of state-of-the-art methods and techniques from the fields of mathematics and computer science (including image processing, biomechanical modeling, and optimization) for radiation oncology.

\vspace{2ex}\noindent\textbf{Peter A. N. Bosman} is a senior researcher in the Life Sciences and Health (LSH) research group at the Dutch national research institute for mathematics and computer science (Centrum Wiskunde \& Informatica) and professor of Evolutionary Algorithms (EAs) at Delft University of Technology. His research concerns the design of scalable model-based EAs and their application, primarily in the LSH domain. Peter has (co-)authored over 100 refereed publications, out of which 4 received best paper awards.

\vspace{1ex}



\end{document}